\documentclass{article}

\usepackage[nonatbib, preprint]{neurips_2020}

\usepackage[utf8]{inputenc} 
\usepackage[T1]{fontenc}    
\usepackage{hyperref}       
\usepackage{url}            
\usepackage{booktabs}       
\usepackage{amsfonts}       
\usepackage{nicefrac}       
\usepackage{microtype}      

\usepackage{graphicx}
\graphicspath{{figures/}}
\usepackage{amsmath,amssymb,amsfonts}
\usepackage{subfigure}
\usepackage{textcomp}
\usepackage{multirow}
\usepackage{xcolor}
\usepackage{algorithm}
\usepackage{algpseudocode}
\usepackage{paralist}
\DeclareMathOperator*{\argmax}{arg\,max}

\title{MEME: Generating RNN Model Explanations via Model Extraction}

\author{
  Dmitry Kazhdan, Botty Dimanov, Mateja Jamnik, Pietro Li\`{o}  \\
  Department of Computer Science and Technology \\
  The University of Cambridge \\
  \texttt{\{dk525,btd26\}@cam.ac.uk }
}

\begin{document}

\maketitle

\begin{abstract}
Recurrent Neural Networks (RNNs) have achieved remarkable performance on a range of tasks. A key step to further empowering RNN-based approaches is improving their explainability and interpretability. In this work we present \textsc{MEME}: a model extraction approach capable of approximating RNNs with interpretable models represented by human-understandable concepts and their interactions. We demonstrate how MEME can be applied to two multivariate, continuous data case studies: Room Occupation Prediction, and In-Hospital Mortality Prediction. Using these case-studies, we show how our extracted models can be used to interpret RNNs both locally and globally, by approximating RNN decision-making via interpretable concept interactions.
\end{abstract}

\section{Introduction} \label{introduction}

Deep Learning (DL) has achieved groundbreaking results in a wide range of fields \cite{zhang2018survey}, and is currently an active research area of Artificial Intelligence (AI). Recurrent Neural Networks (RNNs) are DL models tailored for processing time-series data, and have been successfully applied in domains relying on temporal data processing, including healthcare \cite{miotto2018deep}, Natural Language Processing (NLP) \cite{tarwani2017survey}, video processing \cite{gao2017video}, and many others \cite{fawaz2019deep}.

A central challenge faced by most DL approaches is their lack of interpretability: the black-box nature of most DL models makes it very difficult to directly understand their behaviour. This often leads to a lack of trust in such models. As a result, it is challenging to apply and regulate these models in safety-critical applications, such as healthcare. This lack of interpretability also makes it challenging to extract knowledge from such models, which prohibits users from better understanding their corresponding tasks/domains. As a result, there has recently been a dramatic increase in research on Explainable AI (XAI), focusing on improving explainability/interpretability of DL systems \cite{molnar2019interpretable}.

In particular, one type of XAI approach is \textit{model extraction} (also referred to as \textit{model translation}) \cite{bastani2017interpreting}. Model extraction techniques use rules~\cite{chen2017enhancing}, decision trees~\cite{sato2001rule}, or other more readily explainable models~\cite{kazhdan2020marleme} to approximate complex models, in order to study their behaviour. Intuitively, statistical properties of a complex model should be reflected in an extracted model, provided approximation quality (referred to as \textit{fidelity}) is high enough.

In this work, we present \textsc{MEME}: an approach to RNN (M)odel (E)xplanation via (M)odel (E)xtraction.\footnote{\textit{Meme} refers to a system's behaviour/knowledge passed from one individual to another by imitation, akin to how an extracted model extracts knowledge from an original one via approximation. https://en.wikipedia.org/wiki/Meme} Using \textsc{MEME}, we approximate RNN models with models represented via human-understandable concepts and their interactions. A diagrammatic summary of our approach is given in Figure \ref{vis_abs_fig} and Figure \ref{extr_pred_fig}, and will be discussed in more detail in Section \ref{methodology}. All relevant code is publicly available, and can be found in our open-source repository.\footnote{https://github.com/dmitrykazhdan/MEME-RNN-XAI}

To summarize, we make the following contributions in this work:

\begin{compactitem}
    \item{Present a novel model extraction approach (\textsc{MEME}), capable of approximating RNN models with interpretable models represented by human-understandable concepts and their interactions. To the best of our knowledge, this is the first time concept extraction approaches have been used with RNNs and time-series tasks.}
    
    \item{Quantitatively evaluate \textsc{MEME} using two \textit{multivariate, continuous} data tasks (including a healthcare scenario), showing that \textsc{MEME} generates high-fidelity extracted models. To the best of our knowledge, this is the first time RNN model extraction approaches have been applied to multivariate, continuous data tasks.}
    
    \item{Qualitatively evaluate \textsc{MEME} using the two case studies, showing how extracted models produced by \textsc{MEME} can be inspected in order to achieve both \textit{global} explainability (i.e., explaining overall model behaviour), and \textit{local} explainability (explaining individual predictions) of RNN models.}
    
\end{compactitem}



\section{Related Work} \label{rel_work}


\paragraph{Concept Extraction}    
Most existing works on post-hoc DL explanation methods (those focusing on explaining trained DL models) provide explanations by estimating the importance of input features, and typically focus on local explanations \cite{zeiler2014visualizing}. However, these methods have been shown to have methodological and theoretical limitations \cite{dimanov2020you, adebayo2018sanity}. More recent approaches focus on \textit{concept-based} explanations, presenting explanations in terms of human-understandable \textit{concepts}, which are typically derived from the hidden layers of a DL model \cite{ghorbani2019towards}.
Crucially, existing concept extraction approaches focus predominantly on image recognition tasks, as opposed to time-series tasks, which is the focus of this work. Further details regarding concept-based explanations can be found in Appendix \ref{background}.

\paragraph{RNN Model Extraction}
Several recent works have presented approaches to RNN model extraction, focusing predominantly on language-based modelling NLP tasks. For instance, work in \cite{hou2020learning, weiss2019learning, du2019deepstellar, krakovna2016increasing} presents approaches for approximating RNNs via Finite State Automata (FSA), Probabilistic Deterministic Finite Automata (PDFA), Discrete-Time Markov Chains (DTMC), or Hidden Markov Models (HMMs). Similarly to our work, these works generate extracted models by quantising the RNN hidden space, and approximating the RNN transition dynamics in this space.  However, these works focus on tasks consisting of categorical, univariate data, whereas we consider tasks with multivariate continuous data. Furthermore, we associate the quantised states with human-understandable concepts, as will be discussed in Section~\ref{methodology}.
    
\paragraph{RNN Explainability}
Another direction for RNN explainability focuses on using specialised model architectures \cite{choi2016retain, bai2018interpretable}, or specialised model training procedures \cite{wu2018beyond}. Predominantly, these approaches rely on \textit{attention mechanisms} \cite{choi2016retain, bai2018interpretable}, which require modification of the model architecture and/or model training procedure. In contrast, our approach is agnostic to the RNN architecture, and requires no modifications/retraining. Furthermore, it has recently been argued that ``attention is not explanation'', emphasising that in practice, it is often unclear what relationship exists between attention weights and model outputs \cite{jain2019attention}.


\section{Methodology} \label{methodology}


\subsection{Setup} \label{mth_setup}

Without loss of generality, we consider pre-trained RNN models containing a recurrent layer $l$ with $m$ neurones, processing sequences of inputs $\mathbf{x}_{t}, \mathbf{x}_{t+1}, ...$, with multivariate, continuous input samples $\mathbf{x}_{t} \in \mathbb{R}^n$. By definition, recurrent layers retain a hidden state. Thus, an input inference and its layer $l$ activations at a given timestep can be written as a function $f_{hidden} : \mathbb{R}^n \times \mathbb{R}^m \to \mathbb{R}^m$, such that the function $f_{hidden}$ takes the next input data-point $\mathbf{x}_{t} \in \mathbb{R}^n$, the previous hidden state $\mathbf{h}_{t-1} \in \mathbb{R}^m$, and outputs the next hidden state $\mathbf{h}_{t} \in \mathbb{R}^m$. In this work, we focus on binary classification tasks, in which the model outputs predicted labels $ y \in \{ 0, 1 \} $. Furthermore, we assume that we can access/generate predicted labels at all timesteps, even if the task itself is defined as a whole-sequence prediction task.

\subsection{Concept Extraction} \label{group_extr}

\begin{figure*}[t]
\centering
\includegraphics[scale=0.35]{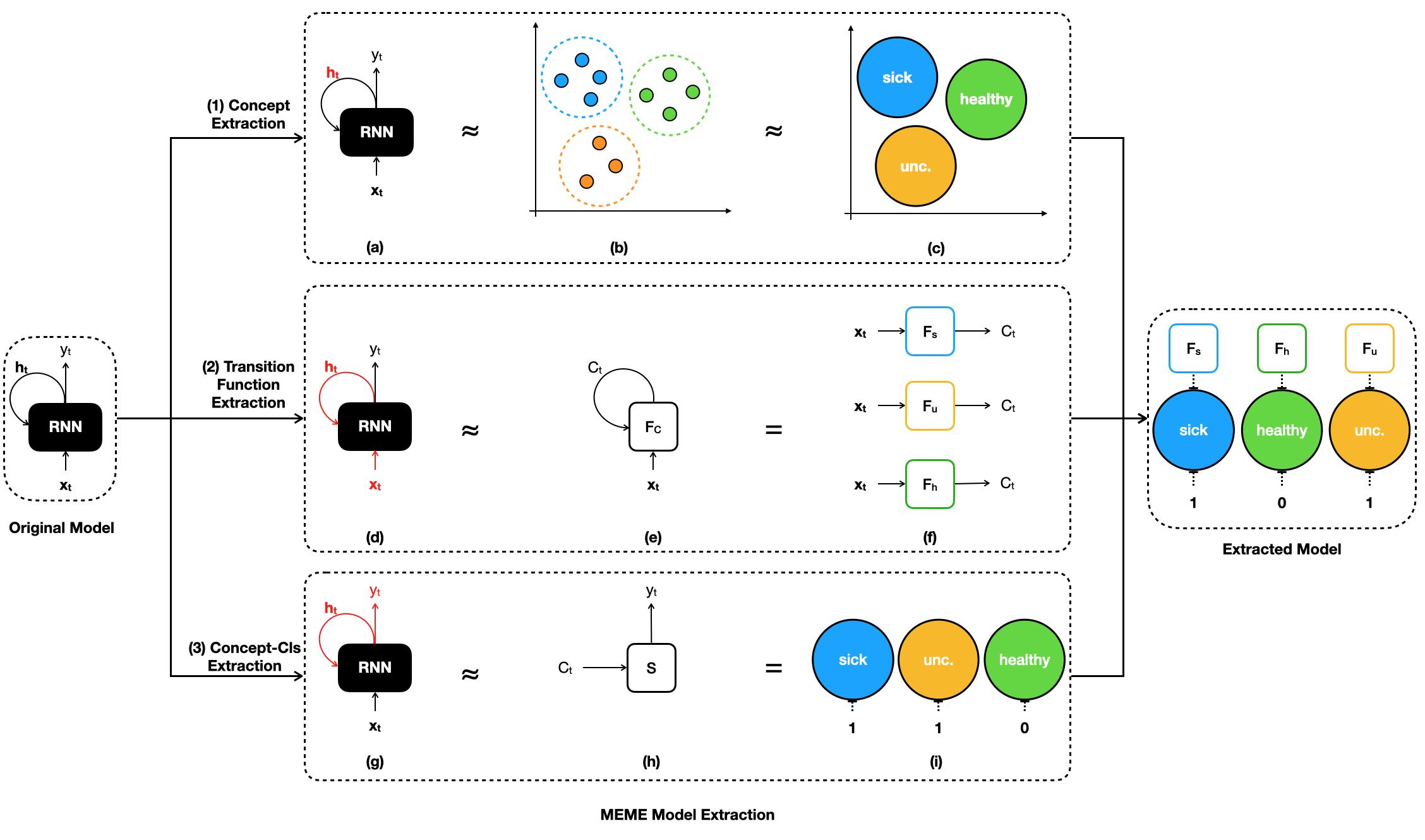} 
\caption{Given an RNN model, we: (1) approximate its hidden space by a set of concepts. (2) approximate its hidden space dynamics by a set of transition functions, one per concept. (3) approximate its output behaviour by a concept-class mapping, specifying an output class label for every concept. For every step in (1)-(3), the parts of the RNN being approximated are highlighted in red. In (a)-(c) we cluster the RNN's training data points in their hidden representation (assumed to be two-dimensional, in this example), and use the clustering to produce a set of concepts (in this case: \textit{sick}, \textit{healthy} and \textit{uncertain}, written as \textit{unc.}). In (d)-(f) we approximate the hidden function of the RNN by a function $F_{C}$, which predicts transitions between the concepts. We represent this function by a set of functions, one per concept (in this case: $F_{s}$, $F_{u}$, $F_{h}$). In (g)-(i) we approximate the output behaviour of the RNN by a function $S$, which predicts the output class from a concept. This function is represented by a concept-class mapping, specifying an output label for every concept (in this case: \textit{healthy} $\to$ 0, \textit{sick} $\to$ 1, and \textit{unc} $\to$ 1). Collectively, steps (1)-(3) are used to produce our extracted model, consisting of concepts, their interactions, and their corresponding class labels.}
\label{vis_abs_fig}
\end{figure*}

The first step of our approach approximates the hidden space of a given RNN with a set of concepts $C$, as shown in Figure~\ref{vis_abs_fig} (a)-(c). 

Similarly to existing concept extraction approaches (such as \cite{ghorbani2019towards}), we extract concepts by obtaining summaries of clusters in the hidden space. We use $f_{hidden}$ to compute the hidden representation of the RNN's training data points, and cluster it using \textit{Kmeans} clustering to generate the set of concepts $C$, such that every cluster is associated with a concept in $C$.

Unlike existing work on concept extraction, which focuses predominantly on image-based tasks, we focus on multivariate, continuous, temporal data tasks. Consequently, we argue that explaining the extracted clusters via prototypical examples is not suitable in our context, and that higher-level explanations should be provided instead. In this work, we rely on the class label space when generating concepts, by introducing a majority-voting cluster naming strategy we refer to as the \textit{majority labelling} approach, which will be described in the remainder of this section.

Given a cluster $k$ consisting of $p$ data-points in their hidden representation $(\mathbf{h}_{1}, ..., \mathbf{h}_{p}) $, and the corresponding class labels of these data-points $(y_{1}, ..., y_{p})$, we define the \textit{majority label} ($MaL$) and the \textit{majority label ratio} ($MaLR$) for that cluster as the most frequently-occurring class label, and it's corresponding ratio of occurrence, respectively, as shown in Equation \ref{mal_eqn} and Equation \ref{malr_eqn}. In Equation \ref{mal_eqn}, $L$ refers to the class label space (thus, in this paper $L = \{ 0, 1 \}$).

\begin{equation} \label{mal_eqn}
MaL_{k} = \argmax_{l \in L} \sum_{i = 1}^{p}[y_{i} = l]  
\end{equation}

\begin{equation} \label{malr_eqn}
MaLR_{k} = \frac{1}{p} \sum_{i = 1}^{p}[y_{i} = MaL_{k}] 
\end{equation}

Given $MaL_{k}$ and $MaLR_{k}$ for a cluster $k$, we define its associated concept $c_{k}$ as shown in Equation \ref{conc_eqn}, where $ClassName(MaL_{k})$ refers to the class name associated with the class label.
\begin{equation} \label{conc_eqn}
c_{k} = ClassName(MaL_{k}) \ \ \textbf{if}\ MaLR_{k} > \theta \ \textbf{,} \ \  unc. \ \textbf{otherwise}
\end{equation}

Hence, the concept information associated with a cluster is the name of the most-frequently occurring class in that cluster, or \textit{uncertain} (written as \textit{unc.} for short), if the $MaLR$ is less than a predefined threshold $\theta$.


\subsection{Transition Function Extraction}

The second step of our approach approximates the RNN's transitions between the extracted concepts, as shown in Figure~\ref{vis_abs_fig} (d)-(f). We achieve this by approximating the RNN's hidden function $f_{hidden}$ with the function $F_{C} : \mathbb{R}^n \times C \to C$ representing the RNN's transition dynamics between concepts. In practice, $F_{C}$ needs to be interpretable (in order to facilitate consequent interpretation of the extracted model), whilst faithfully approximating the original function $f_{hidden}$ (to ensure high fidelity). To meet these criteria, we define $F_{C}$ as a set of interpretable functions, one per concept: $F_{C} = \{ F_{c} \ | \ c \in C \}$. Each function $F_{c}: \mathbb{R}^n \to C $ captures the transition dynamics for a concept $c$. That is, given an input $\mathbf{x}_{t-1}$, $F_{c}(\mathbf{x}_{t-1})$ predicts the concept the RNN will be in at time $t$, assuming the RNN was in concept $c$ at time $t-1$. Crucially, the overall complexity of $f_{hidden}$ is now decomposed into several simpler functions, which can be represented by less complex, more interpretable functions, such as Decision Trees (DTs).

We treat the functions in $F_{C}$ as classifiers, mapping an input $\mathbf{x} \in \mathbb{R}^n$ to a concept $c \in C$. Consequently, for training, these functions are setup as multi-label classification problems, where the functions are equipped with labelled data consisting of input points $\mathbf{x}$, and corresponding concept labels $C$. We obtain the transition function training data directly from the original RNN training data. Further details regarding \textsc{MEME} transition functions can be found in Appendix~\ref{training_data_extr} and Appendix \ref{mth_context_window}.

\subsection{\textsc{MEME} Extracted Models}

The third model extraction step consists of producing a mapping from concepts to output classes, which we refer to as $S : C \to L$, as shown in Figure~\ref{vis_abs_fig} (g)-(i). We define the output class corresponding to a concept as the majority label of its corresponding cluster $MaL$. 

Finally, we define models extracted by \textsc{MEME} as a tuple $(C, F_{C}, S, c_{0})$. We assume that the RNN has a fixed initial hidden state $\mathbf{h}_{0}$, which is located in a cluster corresponding to $c_{0}$ (i.e., the starting concept). Similarly to an FSA, a \textsc{MEME} model processes an input sequence by performing a series of state transitions, such that at timestep $t$, the model computes the next state from the current state and the next input sample, and outputs a class label based on this next state. However, unlike FSA-based approaches, \textsc{MEME} models represent states by interpretable concepts $C$, and represent transition functions by classifiers in $F_{C}$, capable of processing multivariate, continuous input. The sequence classification process of \textsc{MEME} models is summarised in Figure~\ref{extr_pred_fig}. The sequence classification algorithm used by \textsc{MEME} models is given in Appendix \ref{predSeqAlg}.

\begin{figure}[t]
\centering
\includegraphics[scale=0.4]{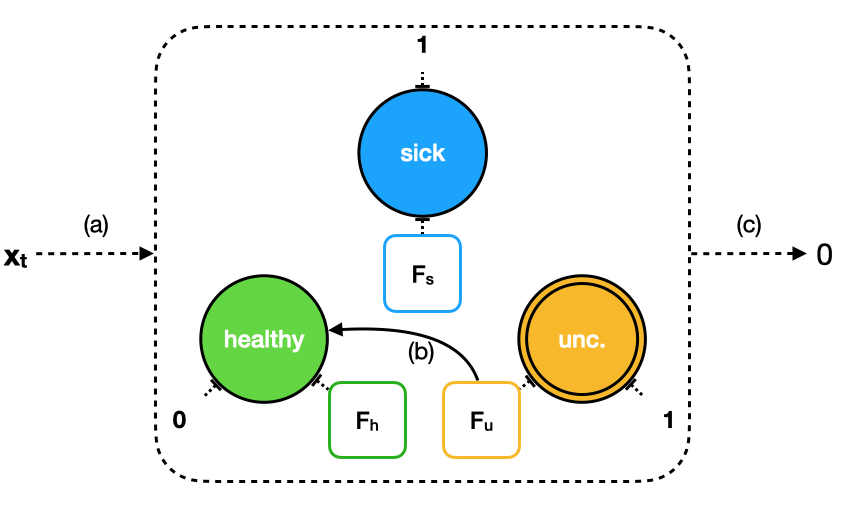} 
\caption{\textsc{MEME} model input processing example. The concept $c_{t-1}$ which the model is in at time $t-1$ is highlighted with a double border. (a) At time $t$, the model receives the next input $\mathbf{x}_{t}$. (b) After receiving the input, the model uses $F_{c_{t-1}}$ (in this case: $F_{u}$) to compute the next concept $c_{t}$ (in this case: \textit{healthy}), from the current concept $c_{t-1}$ (in this case: \textit{unc}) and the new input $\mathbf{x}_{t}$. (c) The model outputs the label corresponding to the next concept $c_{t}$ (in this case: $0$), as the output class label of the input $\mathbf{x}_{t}$.}
\label{extr_pred_fig}
\end{figure} %


\section{Experiments} \label{experiments}

We evaluated our approach using two case-studies: Room Occupation Prediction (ROP), and In-Hospital Mortality Prediction (IHMP). When extracting concepts, we used the RNN recurrent layers closest to the output as the layers computing $f_{hidden}$, for both case-studies. We experimented with using Decision Trees (DTs), and Multi-Layer Perceptrons (MLPs) as transition function types. Further details regarding extracted model specifications can be found in Appendix~\ref{model_spec}. Relevant details and code used to run the experiments can be found in our repository.\footnote{https://github.com/dmitrykazhdan/MEME-RNN-XAI}


\paragraph{Room Occupation Prediction}

The ROP task focuses on predicting room occupancy using measurements of environmental variables. For this task, we make use of the \textit{Occupancy Detection Data Set} available from the UCI Machine Learning repository.\footnote{https://archive.ics.uci.edu/ml/datasets/Occupancy+Detection+} This is a binary classification problem in which observations of sensor readings (such as Temperature, Humidity, or Light) are used to determine whether a room is occupied or unoccupied \cite{room_occ}. Further details can be found in Appendix \ref{rop_task_details}.

\paragraph{In-Hospital Mortality Prediction} \label{ihmp_desc}

For our second case-study, we used the \textit{in-hospital mortality prediction} task defined in \cite{mort_pred}. This work presents a public repository\footnote{https://github.com/YerevaNN/mimic3-benchmarks/} with 4 clinical benchmark tasks (including IHMP), all derived from the Medical Information Mart for Intensive Care (MIMIC-III) dataset \cite{mimic3}. The IHMP task focuses on predicting patient in-hospital mortality based on the first 48 hours of a patient's ICU stay, using a set of clinical variables monitored over time. We used the data extraction pipeline described in \cite{mort_pred} for producing the IHMP task dataset from the MIMIC-III data, and evaluated MEME on a subset of the data the RNN was more confident in (we intend to explore the low-confidence data case in the future). Further details can be found in Appendix \ref{ihmp_task_details}.


\section{Results} \label{results}

This section presents the results obtained by evaluating our approach using the two case studies described above. Section~\ref{results_state} shows the extracted concepts, and the effect of varying concept granularity. Section~\ref{results_transitions} performs global inspection of the transition functions. Section \ref{results_sequence} demonstrates how the extracted models perform sequence prediction, offering local explanations. Section~\ref{results_performance} measures the fidelity of the extracted models, and their performance on the original tasks. Further experimental results can be found in Appendix \ref{further_results}.

\subsection{Concept Granularity} \label{results_state}

\begin{figure}[t]
\centering     
\includegraphics[scale=0.4]{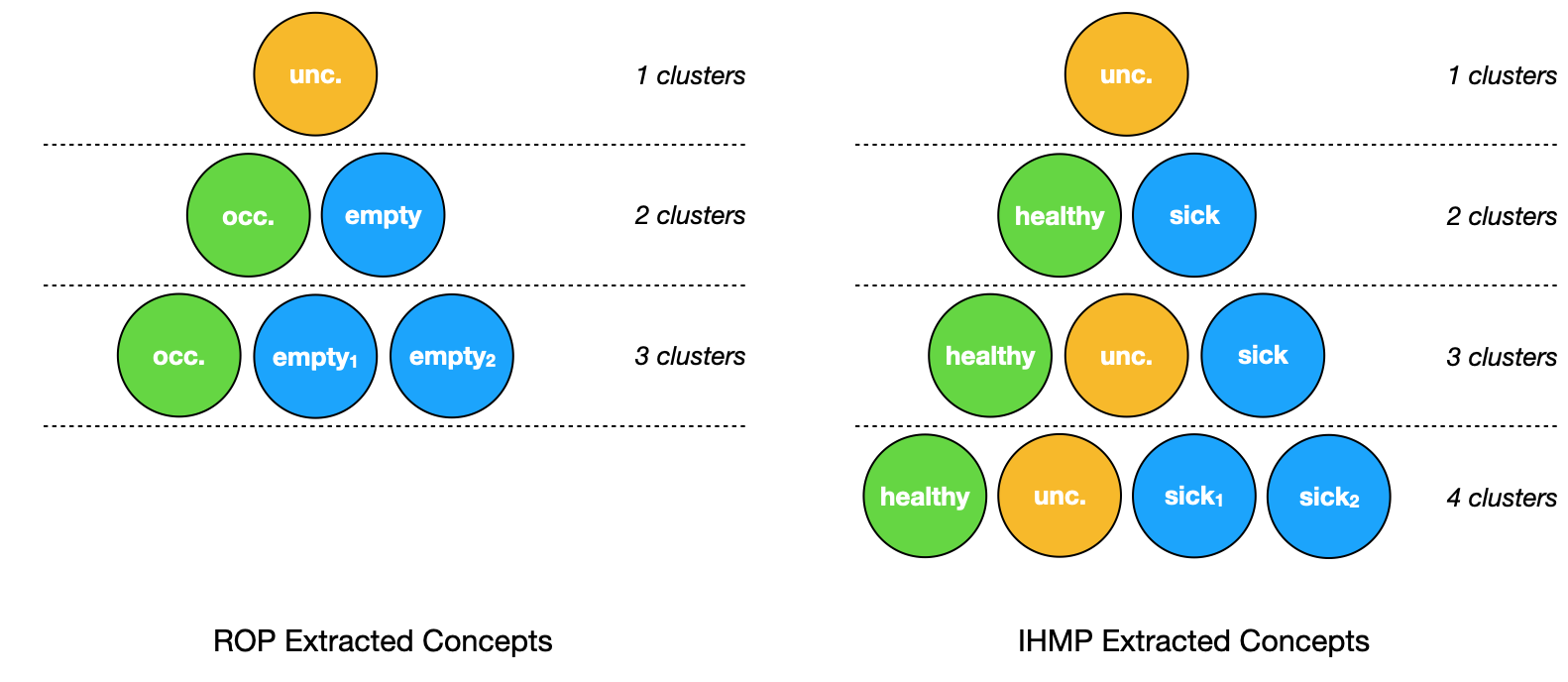}

\caption{The concepts extracted from the RNNs at different clustering levels. Larger cluster numbers exhibited even more repeating concepts than those presented here, and are thus not shown.} \label{state_var}
\end{figure}

We inspect the extracted concepts, and observe how the number of \textit{Kmeans} clusters affects their granularity. Figure~\ref{state_var} shows the extracted concepts obtained at varying numbers of clusters.

A coarse clustering of one cluster yields the \textit{uncertain} concept for both the ROP and the IHMP tasks, since a single cluster consists of a mixture of points from both positive and negative classes. At two clusters, the concepts correspond to the positive and negative class labels. This is expected, since linear separability of the hidden space of a DL model w.r.t. its output classes increases with depth \cite{alain2016understanding}, and is high in layers close to the output. At 3 clusters, the IHMP task produces an extra \textit{uncertain} concept, indicating that the RNN hidden space contains a sub-region corresponding to states in which the RNN is uncertain. At higher concept granularity (4 clusters for IHMP, and 3 clusters for ROP), the clustering step begins to return concepts represented by the same concept labels. Given the rich amount of information contained in the RNN hidden space, these more fine-grained clusterings likely hold information on intra-class concepts (e.g., in case of the IHMP task, $sick_{1}$ and $sick_{2}$ may correspond to different types of sickness progression). 

Overall, for a small cluster number, our \textit{majority labelling} approach offers a straightforward way of interpreting the clusters via their relation to the output classes. For the rest of the experiments in this section, we fix the number of clusters to those producing the most granular concepts with distinct names: 2 clusters in the case of ROP, and 3 in the case of IHMP.

\subsection{Transition Function Interpretation} \label{results_transitions}

In this section, we inspect the transition functions of the extracted models, showing how they can be used to better understand global behaviour and concept interplay of the extracted models. For the sake of space, we focus on a representative subset of the extracted model transition functions, instead of presenting all of them.

\begin{figure}[t]
    \centering
    
    \subfigure[]{\includegraphics[scale=0.15]{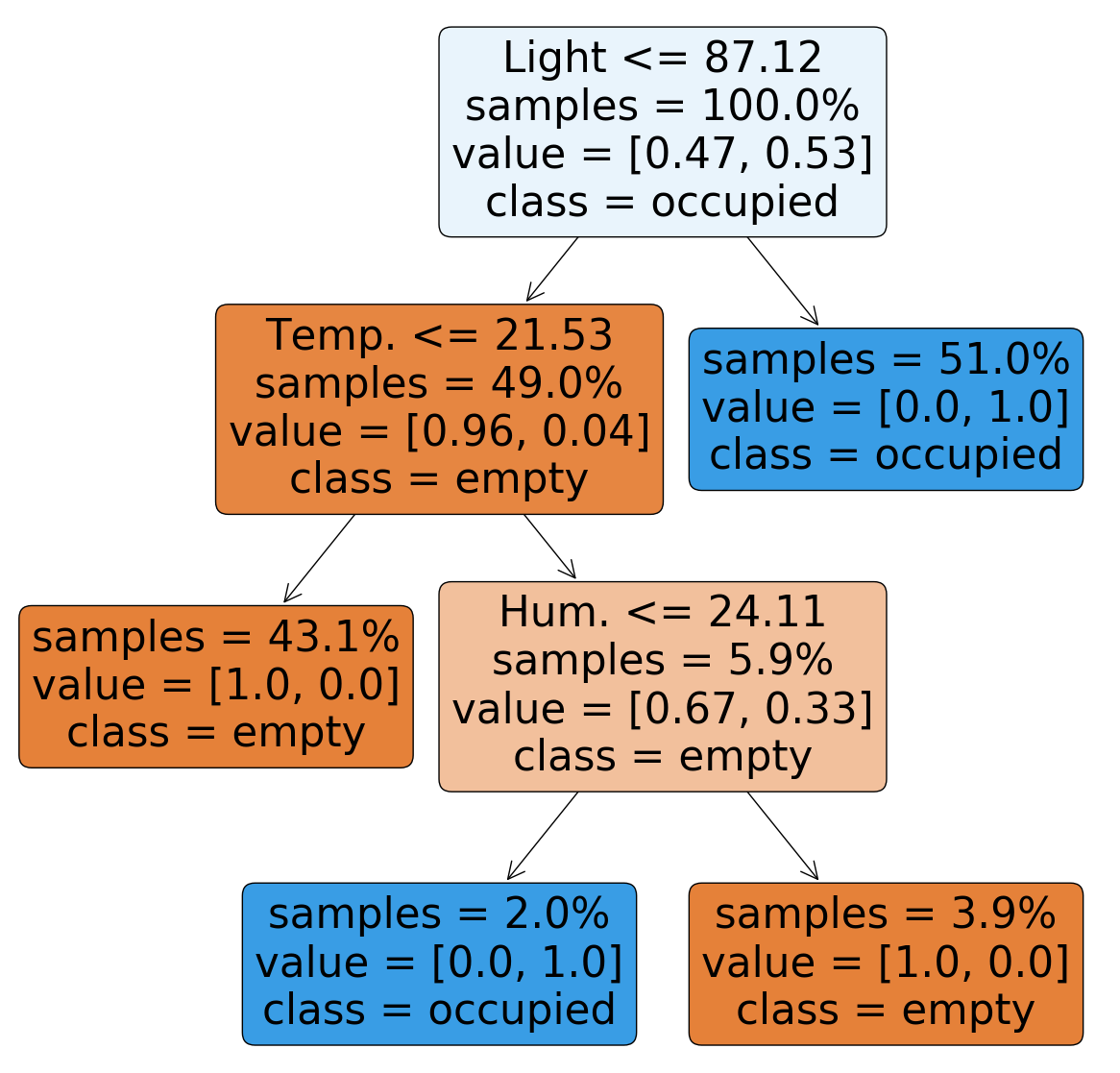}}  \hspace{12mm} 
    \subfigure[]{\includegraphics[scale=0.25]{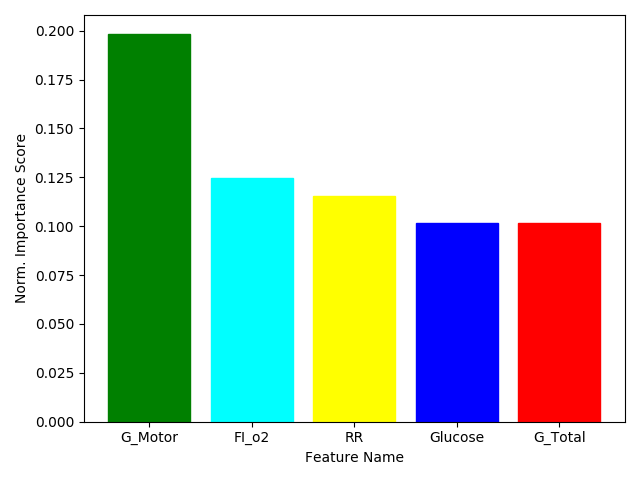}}

    \caption{Transition function interpretation examples. (a) DT transition function for the \textit{occupied} concept of the ROP task. (b) Top 5 most important features and their normalised importance scores for the \textit{uncertain} concept MLP transition function of the IHMP task.}
    \label{trans_fun_fig}
\end{figure}



Since DT transition functions are interpretable, we can inspect them directly. Figure~\ref{trans_fun_fig}(a) shows the DT transition function for the \textit{occupied} concept of the ROP task. Using this plot, it is possible to directly interpret the reasons for the transition function predicting a transition to the \textit{occupied} concept, or to the \textit{empty} concept. In this example, \textit{Light} is the most important feature, with low light indicating that the room is empty. This is likely due to the fact that the first action a room occupant undertakes is to switch on the lights (further details regarding the sensors, room, etc.\ can be found in \cite{room_occ}). The remaining features exhibit similar intuitive justifications. 

In cases where transition functions cannot be interpreted directly (e.g., MLPs), inspection can be done by relying on existing XAI approaches. The field of XAI offers a wide variety of approaches targeted at global explanations of MLPs, including (but not limited to): Partial Dependence Plots (PDPs) \cite{molnar2019interpretable}, model distillation \cite{tan2018learning}, SP-LIME \cite{ribeiro2016should}, GENN \cite{GENN}, and many others \cite{molnar2019interpretable}. The results of one such approach are shown in Figure~\ref{trans_fun_fig}(b), in which we plot the top 5 most important features of the MLP transition function for the \textit{uncertain} concept, using the \textit{permutation feature importance} scoring approach \cite{altmann2010permutation}. Consequently, we are able to better understand the transition behaviour of the MLP function of that concept, and the features it mostly relies on.

\subsection{Local Interpretability} \label{results_sequence}

\begin{figure*}[t]
\centering
\includegraphics[scale=0.35]{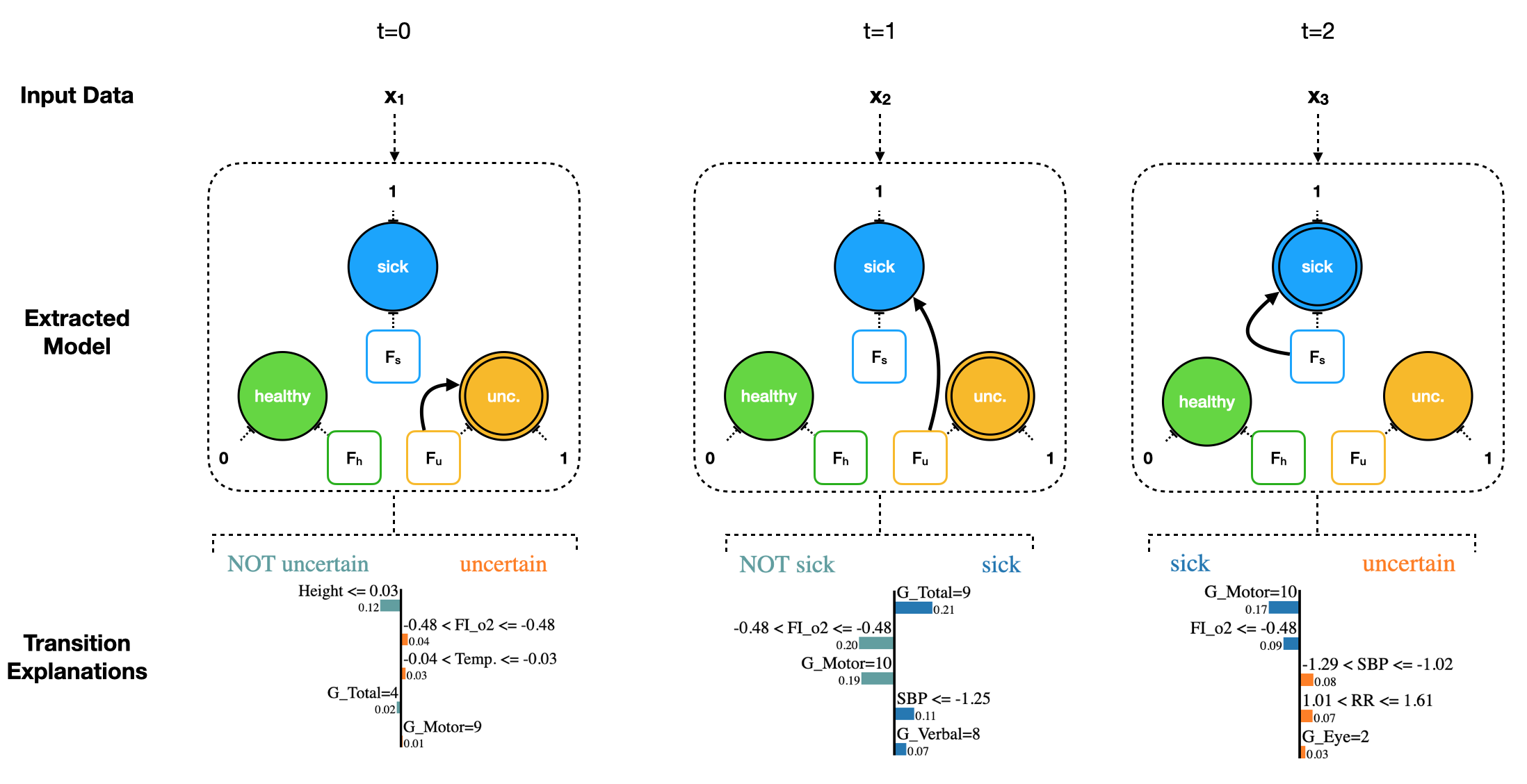} 
\caption{Extracted model sequence processing for three timesteps ($t = 0, 1, 2$), with \textit{uncertain} as the initial concept. For each timestep $t$, the concept the model is in at time $t$ is highlighted with a double border. We show the input data ($\mathbf{x}_{1}$, $\mathbf{x}_{2}$, $\mathbf{x}_{3}$), the corresponding concept transition sequence (\textit{uncertain} $\to$ \textit{uncertain} $\to$ \textit{sick} $\to$ \textit{sick}), and the explanations for each transition function prediction. In this example, the class labels outputted by the model are not shown.}
\label{ihmp_local_exp}
\end{figure*}

In addition to global interpretability, we can obtain local interpretability of an RNN's individual instance prediction by observing sequence processing of the extracted model. Due to how our extracted models are defined, any label produced by such a model at any timestep can be directly traced back to its concepts and their transitions. In case of interpretable transition functions, such as DTs, transition function predictions can be explained directly. In case of black-box transition functions, such as MLPs, we can employ existing methods for producing local explanations of black-box classifier predictions \cite{molnar2019interpretable}. For instance, Figure~\ref{ihmp_local_exp} shows how the MLP-based extracted model of the IHMP task processes a sequence of 3 timesteps, outputting local explanations of the MLP transition function predictions for each timestep. Here, we use the LIME approach \cite{ribeiro2016should} (tabular explainer with discretised continuous values). 


%

\subsection{Model Performance} \label{results_performance}

The utility of our extracted models was also evaluated quantitatively by measuring their performance on the original classification tasks, and their closeness of approximation of the original models. The predictive accuracy of the original RNNs and the extracted models is shown in Table~\ref{acc_perf_tbl}. For the simpler ROP task, both extracted model types achieved performance close to that of the original model. For the relatively more complex IHMP prediction task, extracted models exhibited a slight reduction in performance. However, this reduction was smaller for the MLP-based model, which is likely a consequence of MLPs being more powerful models than DTs \cite{bengio2010decision}.

\begin{table}[t]
\caption{Accuracy Comparison}
\begin{center}
\begin{tabular}{cccc}

	\hline 
	&&\textbf{ROP}&\textbf{IHMP}\\ 
	\hline
	
	\textbf{Original Model}&&                       96.1 +/- 0.1\%&     85.5 +/- 1.1\%\\
	\multirow{2}{*}{\textbf{Extracted Models}}&DT&  95.8 +/- 0.3\%&     79.7 +/- 1.6\%\\
	&MLP&                                           96.0 +/- 0.3\%&     82.3 +/- 0.9\%\\
	
	\hline
	
\end{tabular}
\end{center}
\label{acc_perf_tbl}
\end{table}

\begin{table}[t]
\caption{Approximation Quality Comparison}
\begin{center}
\begin{tabular}{cccc}

	\hline 
	&&\textbf{ROP}&\textbf{IHMP}\\ 
	\hline
	
	\multirow{2}{*}{\textbf{Fidelity}}  &DT     &95.9 +/- 0.3\%    &87.5 +/- 1.0\%\\
	                                    &MLP    &97.1 +/- 0.3\%    & 91.7 +/- 0.8\%\\ \hline
	\multirow{2}{*}{\textbf{Precision}} &DT     &0.92 +/- 0.02     & 0.96 +/- 0.01\\
	                                    &MLP    &0.93 +/- 0.01     & 0.92 +/- 0.01\\ \hline
	\multirow{2}{*}{\textbf{Recall}}    &DT     &0.98 +/- 0.03     & 0.81 +/- 0.02\\
	                                    &MLP    &0.99 +/- 0.01     & 0.93 +/- 0.01\\ \hline
	\multirow{2}{*}{\textbf{F1}}        &DT     &0.95 +/- 0.01     & 0.88 +/- 0.01\\
	                                    &MLP    &0.96 +/- 0.01     & 0.92 +/- 0.01\\
	
	\hline
	
\end{tabular}
\end{center}
\label{fid_perf_tbl}
\end{table}

Table~\ref{fid_perf_tbl} shows the fidelity, precision, recall, and F1 scores of the extracted models, computed by treating the original RNN predictions as the ground truth. Overall, both extracted model types achieved high approximation scores on both tasks. Similarly to the results in Table \ref{acc_perf_tbl}, the MLP-based extracted models achieved a higher overall approximation (fidelity and F1 scores), than the DT-based extracted models, with the difference being more pronounced in the case of the more challenging IHMP task.

These results demonstrate that our approach generates models which are faithful approximations of the original RNNs. Furthermore, they show that for more complex tasks, using more powerful transition functions leads to better performance.


\section{Conclusions} \label{conclusions}

We presented \textsc{MEME}, a model extraction approach capable of approximating RNNs with interpretable models represented by human-understandable concepts and their interactions. We evaluated our approach using two multivariate, continuous data case studies (including a realistic healthcare scenario). Using these case-studies, we demonstrated that extracted models produced by \textsc{MEME} faithfully approximate their original RNN models, and can be used to extract useful knowledge from them.

\textsc{MEME} is a highly modular approach which can be extended in a variety of ways. In this work, we relied on \textit{Kmeans} clustering and the majority labelling approach when performing concept extraction. In the future, we intend to explore other approaches to concept extraction, such as those based on probabilistic clustering (e.g. mixture modelling \cite{reynolds2009gaussian}), or those based on concept directionality (e.g. \cite{yeh2020completeness}). Furthermore, we intend to explore other transition function types, such as Random Forests, or Dynamic Time-Warping models \cite{xing2010brief}. Investigating the effects these approaches have on fidelity, concept granularity, concept representation, etc.\ is an interesting avenue for further exploration.

Furthermore, we relied on the RNN's training data when performing model extraction. Exploring active learning-based approaches capable of querying the RNN and extracting maximally-informative data-points dynamically (e.g., using approaches based on \cite{craven1996extracting}) will likely increase extracted model performance even further, and is thus another interesting direction for future work.


Overall, given the rapidly increasing interest in time-series DL systems (in the field of healthcare, in particular), we believe our approach can play an important role in enriching such systems with explainability and knowledge extraction capabilities.


\appendix

\section{Concept-based Explanations Further Details} \label{background}

Concept-based explanations are explanations consisting of high-level, human-understandable ``concepts''. In practice, giving a precise definition for what constitutes a concept from a psychological and/or philosophical standpoint is highly non-trivial \cite{ghorbani2019towards}. Instead, in line with recent work, we refer to a concept as a global explanation of the internal workings of a model, presented in higher-level, human-understandable units, rather than individual features, pixels, or characters. Concept-based explanations are intended to serve as more understandable, usable explanations of DL models, and have been used in a range of different ways, including: inspecting what the model has learned \cite{ghorbani2019towards}, providing class-specific explanations \cite{kim2017interpretability}, and discovering causal relations of concepts in data \cite{goyal2019explaining}. Most recent work on concept extraction (extracting learned concepts from DL models) focuses on image-recognition tasks, representing concepts by images (e.g., concepts associated with the class ``dog'' may be images of paws, ears, or tails) \cite{kim2017interpretability}.

Deep Neural Networks (DNNs) have been shown to perform hierarchical feature extraction, with layers closer to the output utilising higher-level data representations, compared to layers closer to the input \cite{hinton2007learning,DBLP:journals/corr/ZhouKLOT14}. Consequently, recent work on concept extraction relies on euclidean distance in the activation space of final layers (referred to as the \textit{hidden space}) of DNN models as an effective concept similarity metric \cite{ghorbani2019towards}. The intuition is that data-points exhibiting similar high-level features will be clustered together in the hidden space of the corresponding DNN layers, implying that the hidden space is expected to demonstrate conceptual grouping. Thus, one approach to concept extraction is to cluster data points in the activation space of higher-level model layers, and extracting cluster summaries in the form of prototypical samples, to serve as concept representations \cite{ghorbani2019towards}.

\begin{figure}[t]
    \centering
    
    \subfigure[Low Granularity]{\includegraphics[scale=0.6]{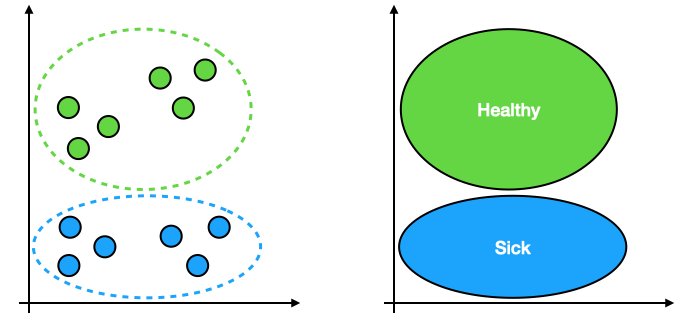}}
    
    \subfigure[High Granularity]{\includegraphics[scale=0.6]{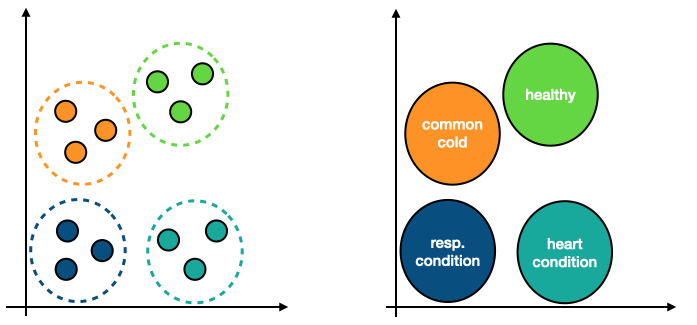}}

    \caption{Anticipated effect of cluster number on concept granularity. The subfigures on the left demonstrate two different \textit{Kmeans} clusterings of the same hidden representation. The subfigures on the right demonstrate the corresponding extracted concepts. (a) A small cluster number yields coarse concepts (\textit{sick} and \textit{healthy}). (b) A higher cluster number increases concept granularity. In this case, extracted concepts now represent specific sicknesses/conditions (\textit{common cold}, \textit{respiratory condition} and \textit{heart condition}).}
    \label{concept_granularity}
\end{figure}

Intuitively, we expect the number of clusters to affect the granularity of the corresponding extracted concepts. Thus, a low cluster number will likely correspond to extraction of coarse concepts, and a high cluster number will likely increase concept granularity. An example is given in Figure~\ref{concept_granularity}.

\section{MEME Model Extraction Further Details}

\subsection{Transition Training Data Extraction} \label{training_data_extr}

For every time-series sequence $X_{s} = (\mathbf{x}_{1}, ..., \mathbf{x}_{T})$ in the RNN training data, we obtain the corresponding RNN hidden states $H_{s} = (\mathbf{h}_{1}, ..., \mathbf{h}_{T})$, and use $H_{s}$ to obtain the corresponding concept sequence $C_{s} = (c_{1}, ..., c_{T})$. We then use $C_{s}$ to produce the sample transition dataset $Tr_{s} = ([c_{0}, \mathbf{x}_1, c_{1}], ..., [c_{T-1}, \mathbf{x}_T, c_{T}])$, such that every item in $Tr_{s}$ consists of the concept at time $t-1$, the input point at time $t$, and the next concept at time $t$, for $t$ ranging from $1$ to $T$. We aggregate these transition samples $Tr_{s}$ over all training sequences, to produce the overall transition dataset $Tr$. Using $Tr$, the training datasets $D_{c}$ for every transition function $F_{c}$ are defined as:
\begin{equation}
D_{c} = \{ (\mathbf{x}_{t}, c_{t}) \ | \ [c_{t-1}, \mathbf{x}_t, c_{t}] \in Tr \ \land \ c_{t-1}=c \} 
\end{equation}

In practice, the nature and sizes of the transition datasets $D_{c}$ are unpredictable. Hence, their important properties (e.g., class imbalance, size, or representativity) are difficult to anticipate/control. For now, we account for one of the mentioned issues (class imbalance) by balancing out every dataset $D_{c}$ (via downsampling), ensuring that they contain an equal number of class samples.

\subsection{Transition Function Context Window} \label{mth_context_window}

\begin{figure}[t]
\centering     

\subfigure[Hidden Space]{\includegraphics[scale=0.4]{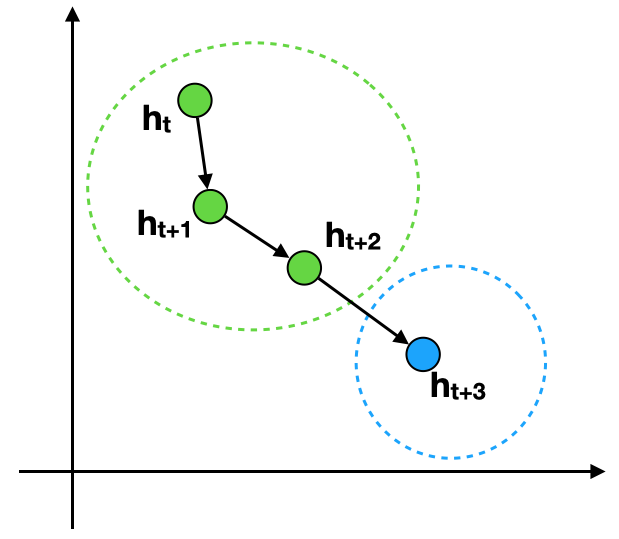}} \hspace{8mm}
\subfigure[Quantised Space]{\includegraphics[scale=0.4]{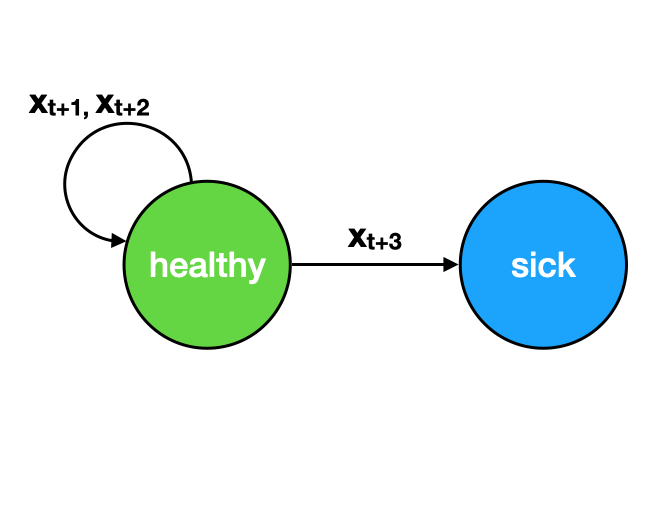}}

\caption{(a) When processing an input sequence ($\mathbf{x}_{t+1}$,$\mathbf{x}_{t+2}$,$\mathbf{x}_{t+3}$), the RNN outputs a corresponding sequence of points ($\mathbf{h}_{t+1}$,$\mathbf{h}_{t+2}$,$\mathbf{h}_{t+3}$) in the hidden space (assumed to be two-dimensional in this example). These points move within a cluster (intra-cluster dynamics), before moving into another cluster (inter-cluster dynamics). (b) Due to quantisation, the intra-cluster movement of datapoints is not observable, implying that the extracted model only observes the final sample ($\mathbf{x}_{t+3}$) causing the model to transition to the other cluster.
Note: in (b), we omit the transition functions and concept-class map here for clarity, showing only the concept sequence.} \label{clust_mvmnt}
\end{figure}

The clustering quantisation step described in Section~\ref{group_extr} discards information about the \textit{intra-cluster dynamics} of an RNN (i.e., hidden space dynamics within clusters), and retains only \textit{inter-cluster dynamics} information, as summarised in Figure~\ref{clust_mvmnt}. In practice, this is a potential source of approximation error, which can adversely affect extracted model fidelity. We address this issue by introducing a \textit{context window} hyper-parameter $w$ to our transition functions, allowing them to retain a context window containing $w$ previous input points. In this case, the transition functions are defined as $F_{c} : \mathbb{R}^{n*(w+1)} \to C$, where $F_{c}$ takes the current input $\mathbf{x}_{t}$, together with $w$ previous inputs $\mathbf{x}_{t-1}, ..., \mathbf{x}_{t-w}$ concatenated together, when predicting the next concept $c_{t}$.

\subsection{Sequence Processing} \label{predSeqAlg}

\begin{algorithm} [h]
\caption{classifySequence($sequence$)} \label{predSeq}
\begin{algorithmic}[1]

\State $c = c_{0}$
\State $labels = []$

\For{$t \in [ w, ..., T ] $}
    \State $label = S(c)$ \Comment{Get next output label}
    \State $labels.append(label)$ 
	\State $input = sequence[(t-w):t]$
	\State $c = F_{c}(input)$  \Comment{Get next concept}
\EndFor

\State \textbf{return} $labels$

\end{algorithmic}
\end{algorithm}

In Algorithm \ref{predSeq}, $w$ is the transition function context window size (assumed to be less than the sequence length $T$), $sequence$ is an input sequence to be classified, and $F_{c}$ denotes the transition function of a concept $c$. In case of full-sequence classification (classifying not every timestep, but the entire sequence), we simply retrieve the last label from $labels$. We have also implemented a more efficient, batched implementation of Algorithm \ref{predSeq} (not shown here for the sake of space), which can be found in our repository.

\section{Experimental Setup Further Details} \label{model_spec}

\subsection{Concept Extraction}
When selecting a value of the threshold $\theta$ used for computing concept names, we found that setting $\theta = 0.8$ works well in practice. Thus, all results reported in Section \ref{results} were computed with $\theta$ fixed to this value. We intend to explore more principled practices for choosing $\theta$ as future work.

\subsection{Transition Functions}
For DTs, we relied on the \textit{DTClassifier} implementation provided in scikit-learn \cite{scikit-learn}. Restricting the maximum depth of the DTs exhibited a beneficial regularising effect, reduced overfitting and improving predictive performance, whilst also making inspection easier. Setting a maximum depth of 2 worked best for the ROP task, and a depth of 3 for the IHMP task. 

For MLPs, we used a 3-layer MLP, consisting of a softmax output layer, and two hidden ReLu dense layers of sizes 200 and 50. We used the \textit{Keras} API available in \textit{Tensorflow}\footnote{https://www.tensorflow.org/guide/keras/overview} for setting up the models, and trained with the standard categorical crossentropy loss. Further details can be found in the repository.

\section{ROP Task Further Details} \label{rop_task_details}

\subsection{Features \& Preprocessing}

The original dataset consists of 5 features: \textit{Temperature} (Temp.), \textit{Humidity} (Hum.), \textit{Light}, \textit{Humidity Ratio}, and \textit{Co2}. We found that the \textit{Humidity Ratio} feature was practically equivalent to the \textit{Humidity} feature (up to a scale factor), and therefore discarded it. Hence, our ROP dataset consisted of 4 features.  

The original dataset consists of three sequences of sensor readings taken at regular $60$s intervals (i.e., the train/test/val datasets each consist of a single long sequence). In order to produce multi-sample datasets, we split each of these 3 sequences into subsequences of $60$ timesteps each. 

\subsection{RNN Model}

The RNN model trained for this task consisted of 2 hidden LSTM layers of sizes 100 and 50, with dropout layers in-between, and a sigmoid output layer. The model was trained with a binary crossentropy loss.

\subsection{Classes}

The problem is formulated as a binary classification problem, with class $0$ corresponding to an empty room (referred to as \textit{empty} in the rest of this work), and class $1$ to an occupied room (referred to as \textit{occupied} in the rest of this work, and abbreviated as \textit{occ.}). The RNN was trained to compute a predicted label at every timestep.

\section{IHMP Task Further Details} \label{ihmp_task_details}

\subsection{Features \& Preprocessing}

For our RNN model, we relied on the data representation used in \cite{mort_pred}. The input features consist of 17 clinical variables collected during the first 48 hours of a patient's stay. Table \ref{ihmp_features} lists the input features, together with their corresponding acronyms, and whether they are continuous or not (referred to as \textit{type}). In addition to the 17 \textit{clinical} variable features, the feature set also included 17 binary features (one per clinical feature), specifying whether the values of the clinical variables were observed, or inferred (these binary features are referred to as the \textit{non-clinical} features). We used this original feature representation without modification when training the RNN model.

\begin{table}[t]
\caption{IHMP Feature Names, Acronyms, and Types}
\centering
\begin{tabular}{| c | c | c |} \hline

\textbf{Name}                       & \textbf{Acronym}      & \textbf{Type}     \\ \hline

Capillary refill rate               & CRR                   & Categorical       \\ \hline
Glascow coma scale eye opening      & G\_Eye                & Categorical       \\ \hline
Glascow coma scale motor response   & G\_Motor              & Categorical       \\ \hline
Glascow coma scale total            & G\_Total              & Categorical       \\ \hline
Glascow coma scale verbal response  & G\_Verbal             & Categorical       \\ \hline

Diastolic blood pressure            & DBP                   & Continuous        \\ \hline
Fraction inspired oxygen            & FI\_o2                & Continuous        \\ \hline
Glucose                             & Glucose               & Continuous        \\ \hline
Heart Rate                          & HR                    & Continuous        \\ \hline
Height                              & Height                & Continuous        \\ \hline
Mean blood pressure                 & MBP                   & Continuous        \\ \hline

Oxygen saturation                   & o2S                   & Continuous        \\ \hline
Respiratory rate                    & RR                    & Continuous        \\ \hline
Systolic blood pressure             & SBP                   & Continuous        \\ \hline
Temperature                         & Temp.                 & Continuous        \\ \hline
Weight                              & Weight                & Continuous        \\ \hline
pH                                  & pH                    & Continuous        \\ \hline

\end{tabular}
\label{ihmp_features}
\end{table}

For the extracted models, we relied on a slightly modified input representation: the categorical features were represented in their categorical format (as opposed to the one-hot encoded format used for the RNN), and the \textit{non-clinical} features were omitted. This configuration gave slightly better extracted model performance, and also made the extracted models easier to interpret.

Before training the RNN, we balanced the dataset via down-sampling, ensuring an equal number of positive and negative class samples. Furthermore, we focused on extracting models encapsulating the RNN's typical behaviour, instead of edge-case behaviour, by only using data the RNN was more confident in. This was achieved by using the RNN's sigmoid output as a proxy for confidence, and selecting ``high-confidence'' points from the balanced dataset, for which this output was either in $[ 0, 0.2 ]$, or in $[0.8, 1.0]$ (i.e., ``closer to 0'', or ``closer to 1''). Extracted model training and evaluation was consequently performed using this ``high-confidence'' data. 

\subsection{Model}

The RNN model trained for this task consisted of 3 hidden LSTM layers of sizes 256, 100 and 100, with dropout layers in-between, and a sigmoid output layer. The model was trained with a binary crossentropy loss.

\subsection{Classes}

As discussed previously in Section~\ref{ihmp_desc}, the task is formulated as a binary classification problem, with class $1$ corresponding to a patient dying in the first 48 hours of ICU stay (referred to as \textit{sick}), and class $0$ to a patient not dying (referred to as \textit{healthy}). The RNN was trained for sequence prediction, computing a predicted label only at the end of an entire sequence.

\section{Further Results} \label{further_results}

\subsection{Transition Function Interpretation} \label{extra_trans_fun}

\begin{figure}[t]
\centering
\includegraphics[scale=0.55]{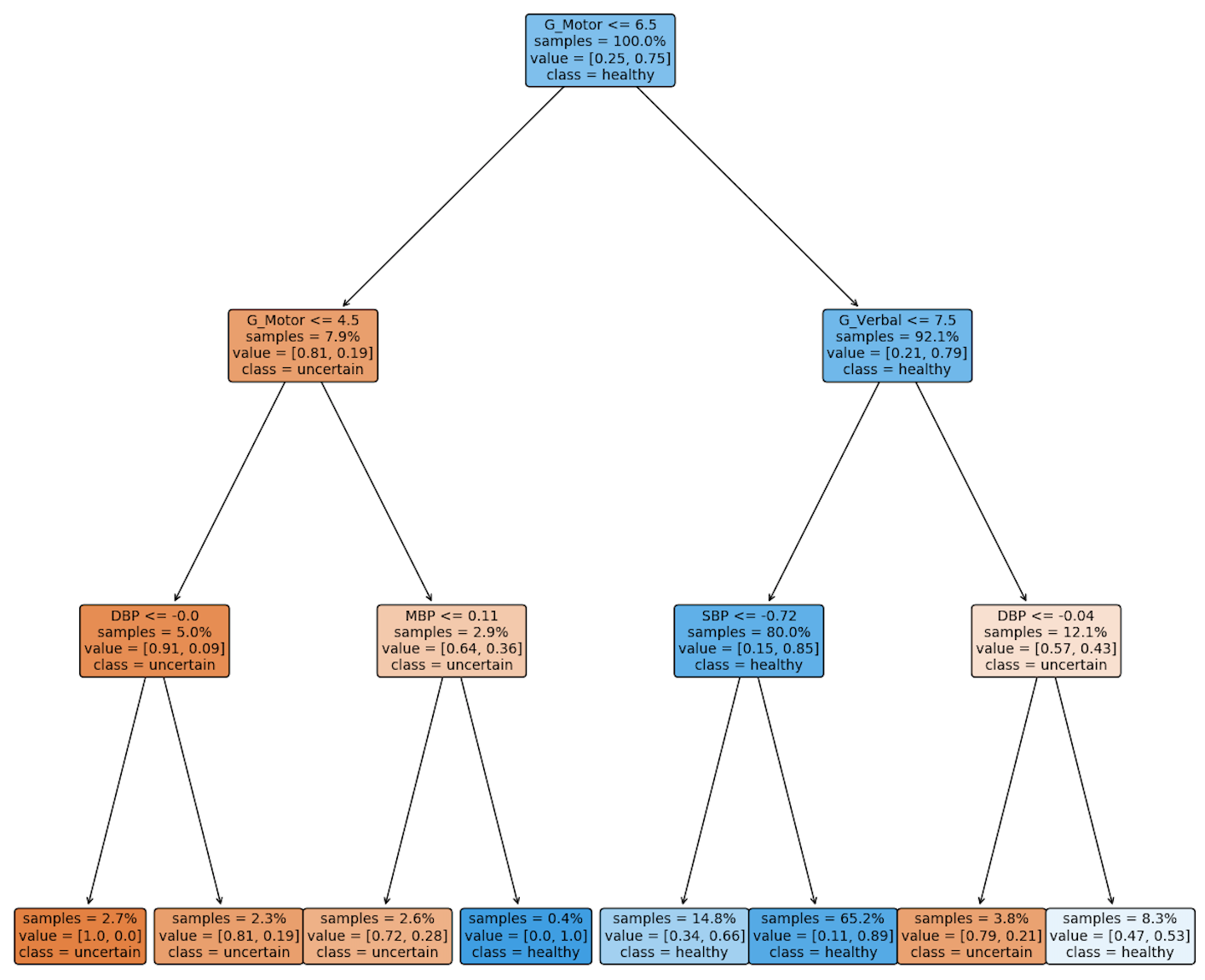} 
\caption{DT transition function for the \textit{healthy} concept of the IHMP task. The original RNN model never exhibited a direct transition from the \textit{healthy} concept to the \textit{sick} concept. Thus, this DT only has two output classes (\textit{uncertain} and \textit{healthy}).}
\label{ihmp_dt}
\end{figure}

Similarly to the ROP task, Figure~\ref{ihmp_dt} shows a DT transition function of the IHMP task. As in the ROP example, these plots can be used to inspect the reasons for the extracted model concept transitions, which could be used to better understand the nature of sickness progression and the key reasons for the model to ``change it's mind'' over time.

\subsection{Transition Function Comparison} \label{extra_fun_comp}

\begin{figure}[t]
\centering
\subfigure[]{\includegraphics[scale=0.25]{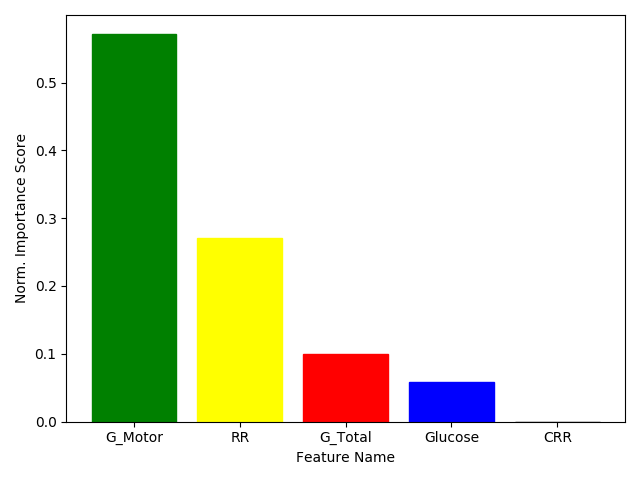}}  \hspace{12mm} 
\subfigure[]{\includegraphics[scale=0.25]{dnn_fnb.png}}

\caption{Top 5 most important features and their normalised importance scores for the \textit{uncertain} concept of the DT(a), and MLP(b) extracted models for the IHMP task.}
\label{fnb_fig}
\end{figure}

Figure \ref{fnb_fig} shows the top 5 most important feature scores of the transition function for the \textit{uncertain} concept, for both the DT and MLP extracted models. The top 4 features of the DT model are all in the top 5 for the corresponding MLP model, indicating that both models were capable of identifying and using the discriminative power of the same features.%
%

Interestingly, the MLP model has the \textit{FI\_O2} feature as the second most important one, which might be due to the MLP capability of relying on more complex, non-linear features, unusable by a DT \cite{bengio2010decision}. More generally, Figure \ref{fnb_fig} demonstrates that it is possible to compare and contrast the results of different transition function types in order to better understand the underlying features and model behaviour.

\subsection{Context Window Size} \label{results_context}

For the experiments described in Section \ref{results}, Appendix \ref{extra_trans_fun}, and Appendix \ref{extra_fun_comp}, we set a context window size of $w=0$ (i.e., the transition functions retain no context). In this section, we quantitatively evaluated the effect of using the transition function context window described in Appendix~\ref{mth_context_window}. Figure~\ref{window_effect} shows the effect of the context window size $w$ on the $F1$ approximation score of the extracted models. %
\begin{figure}[t]
\centering
\includegraphics[scale=0.3]{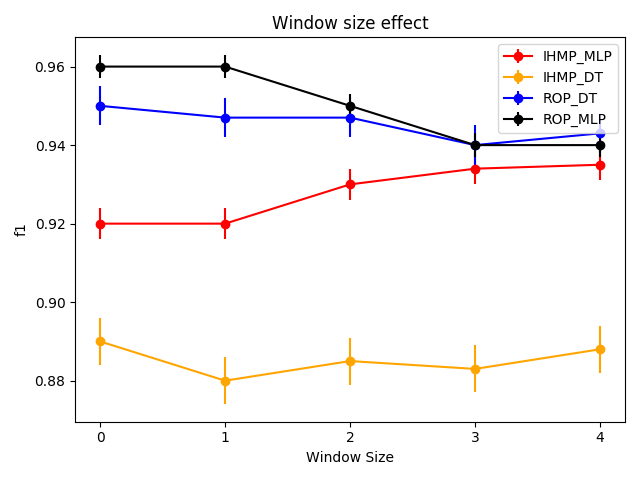} 
\caption{Effect of window size $w$ on the F1 score.}
\label{window_effect}
\end{figure} 

For the ROP task, both the DT and MLP models exhibited a $\sim$1-2\% decrease in performance with increasing window size. The likely reason for this is that the input features at the current timestep are sufficient for good performance in case of simpler tasks, such as ROP. In such cases, a context window introduces extra unnecessary input features to the transition functions, harming their performance, and the overall performance of the extracted model.

For the IHMP task, increasing the window size did not affect the DT-based model score significantly, but resulted in a $\sim$2\% increase in performance for the MLP-based model. The likely reason for this is that, unlike the MLP transition functions, the DT transition functions were not able to extract information signals contained in the context information \cite{bengio2010decision}. 

Overall, our results indicate that for complex tasks (such as IHMP) and for powerful transition functions (such as MLPs), a context window may improve extracted model approximation quality. However, this might not be the case for relatively simpler tasks (such as ROP) or simpler transition functions (such as DTs). In future work, we intend to explore context window size variation in more detail.


\end{document}